\documentclass[11pt]{article}

\usepackage{dialogue}

\usepackage{ifxetex}
\ifxetex
    \usepackage{fontspec}
    \setromanfont{Times New Roman}
\else
  \usepackage{cmap}
  \usepackage[T2A,T1]{fontenc}
  \usepackage[utf8]{inputenc}
  \usepackage{times}
  \usepackage{latexsym}
  \usepackage{substitutefont}
  \substitutefont{T2A}{\familydefault}{cmr}
\fi

\usepackage[russian,british]{babel}
\usepackage{url}
\usepackage{pgf}

\usepackage{covington} %% comment it out if you do not require this option

\usepackage{hyperref}

\dialogfinalcopy % Uncomment this line for the final submission

\title{A new approach to calculating BERTScore for automatic assessment of translation quality}

\author{Vetrov A.A. \\
  NUST “MISIS” \\
  Moscow, Leninskiy prospect, 4 \\
  {\tt andrej.vetrov@edu.misis.ru} \\\And
  Gorn E.A. \\
  NUST “MISIS” \\
  Moscow, Leninskiy prospect, 4 \\
  {\tt gorn.ea@misis.ru} \\}

\date{}

\begin{document}
\maketitle
\begin{abstract}
The study of the applicability of the BERTScore metric was conducted to translation quality assessment at the sentence level for English -> Russian direction. Experiments were performed with a pre-trained Multilingual BERT as well as with a pair of Monolingual BERT models. To align monolingual embeddings, an orthogonal transformation based on anchor tokens was used. It was demonstrated that such transformation helps to prevent mismatching issue and shown that this approach gives better results than using embeddings of the Multilingual model. To improve the token matching process it is proposed to combine all incomplete WorkPiece tokens into meaningful words and use simple averaging of corresponding vectors and to calculate BERTScore based on anchor tokens only. Such modifications allowed us to achieve a better correlation of the model predictions with human judgments. In addition to evaluating machine translation, several versions of human translation were evaluated as well, the problems of this approach were listed.
  

\end{abstract}

\selectlanguage{british}
\section{Introduction}
\label{intro}
%
% The following footnote without marker is needed for the camera-ready
% version of the paper.
% Comment out the instructions (first text) and uncomment the 8 lines
% under "final paper" for your variant of English.
% 

To assess machine translation, which appeared in the middle of the 20th century, a comparison with the reference translation, performed by people, the so-called “gold standard”, was traditionally used. With the improvement of the machine translation quality, and especially with the beginning of the use of deep neural networks since 2014\cite{Kyunghyun_Cho-2014}, it became possible to increase the estimates of translation fluency and adequacy \cite{Andy_Way-2019}. In addition, there have been attempts to identify universal metrics with which you can automatically evaluate the quality of arbitrary translation, including those made by people. 

The principles and methods used to assess the quality of machine translation are described a lot \cite{Lifeng_Han-2021}. Among such methods there are both Human \cite{Markus_Freitagy-2021} and automatic ones. The purpose of the automatic assessment is to obtain some score that is as close as possible to the expert's score but does not require his involvement. 

The existing automatic methods can be divided into 2 groups according to the metrics they use. The first group uses metrics such as: BLUE \cite{Papineni-2002}, METEOR \cite{Satanjeev_Banerjee-2005} and others. All the listed metrics cannot be considered as fully automatic since they require a reference translation. Meanwhile their quality depends on the quality of such translation. The second group, which appeared recently, includes YiSi \cite{Chi_kiu-2019}, BERTScore \cite{Tianyi_Zhang-2019} and its variations. Further in this paper we focused on the study of BERTScore.

\section{BERTScore and mismatching issue}
The metric was first proposed to assess the quality of text generation with comparing candidate sentences to annotated references \cite{Tianyi_Zhang-2019}. For this type of tasks, the metric showed the greatest correlation with the estimates given by humans among other metrics. Metric uses contextual embedding of tokens of the pre-trained BERT model \cite{Devlin-2019}.
\begin{figure}[h]
  \centering
  \includegraphics[angle = 90, trim = 8.3cm 7cm 7cm 7cm, width=15cm]{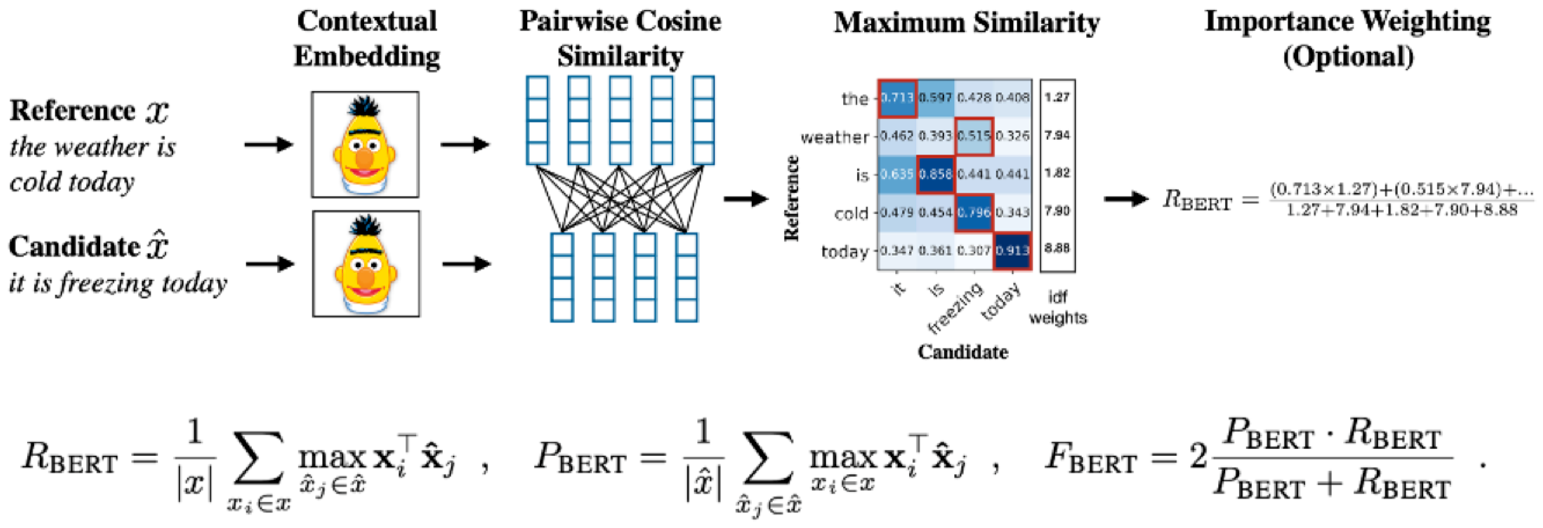}
  \setlength{\belowcaptionskip}{-1cm} \caption{The BERTScore calculation as given in the original article by Zhang et al.}
  \label{bertscore}
\end{figure}
For each token from the candidate sentence, the closest token is determined in the reference sentence, and vice versa, to obtain Recall and Precision, respectively. The obtained values of cosine distances are averaged. Further, Recall and Precision are combined to calculate F1 as shown at Figure~\ref{bertscore}.

Considering the influence of token frequency, the authors suggested using idf-weighting. Impact of this approach depends on the choice of the original document corpus and slightly affected the results.

Hereafter, Lei Zhou, Liang Ding and Koichi Takeda \cite{Lei_Zhou-2020}, as participants in the Quality Estimation competition at the 5th conference on Machine Translation, tried to adopt BERTScore to unsupervised QE without human assessment. They tried the pretrained multilingual BERT and revealed so-called mismatching issue. 

\begin{figure}[h]
  \centering
  \includegraphics[angle = 270, trim = 0cm 0cm 0cm 0cm, width=13cm]{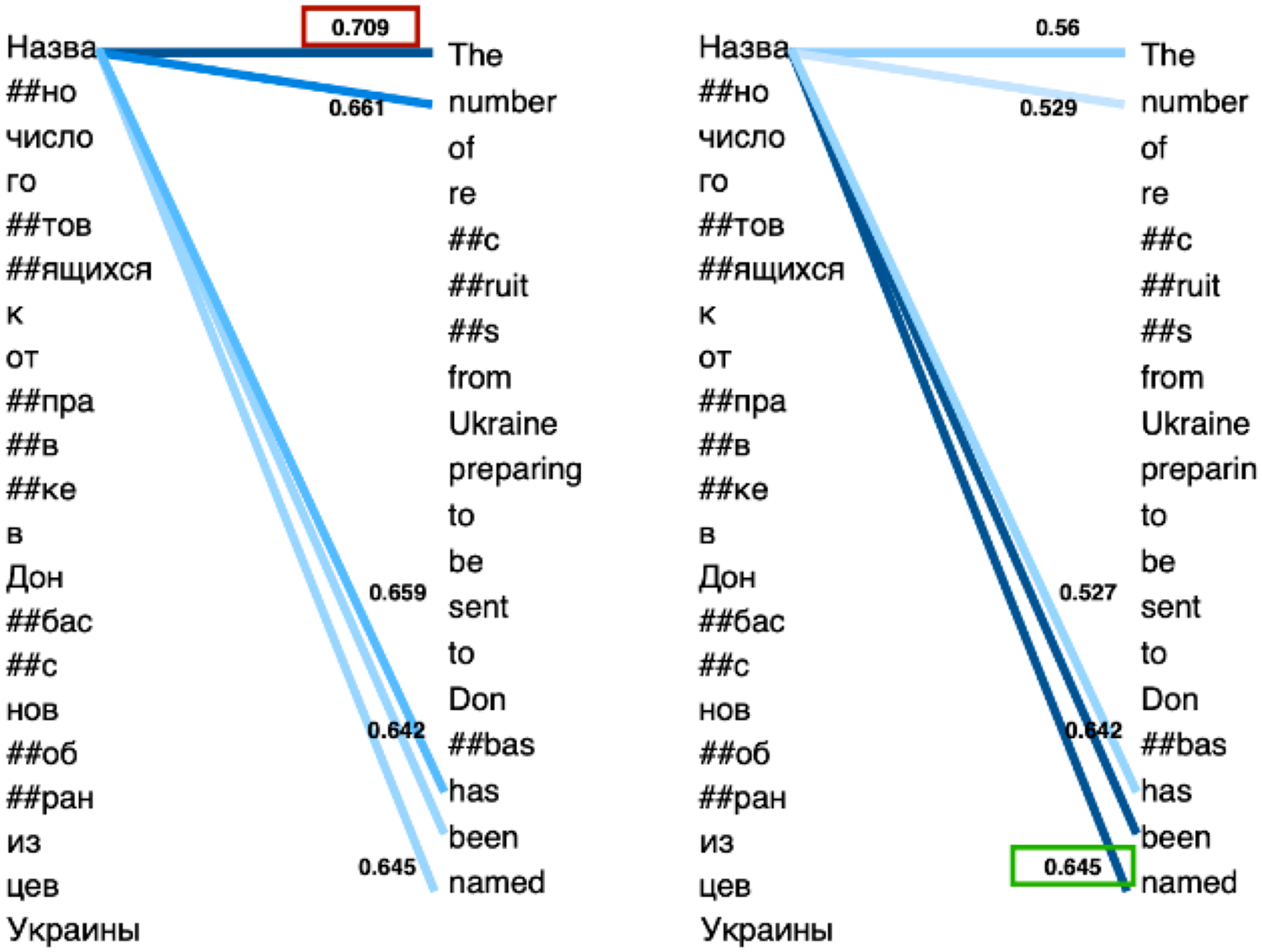}
  \caption{An example of the mismatching issue from the work by Lei Zhou et al., where the Russian token \selectlanguage{russian}"Назва" \selectlanguage{british} is mismatched to the English token "The". After addition weighting authors could improve the matching.}
  \label{mi}
\end{figure}

The problem appeared strongly for the ru->en direction. The reason was the method of tokenization in the model. BERT uses WordPiece, a tokenizer that splits text either into full words or into piece of words, for which embedding vectors are then calculated. Pre-trained Multilingual BERT tokenizer splits ru-words into too small pieces, thus during BERTScore calculation these pieces do not find corresponding en-words in the sentence in English, as in the example from the article Figure~\ref{mi}. 
In response to this issue, the authors \cite{Lei_Zhou-2020} proposed to use additional token alignment. Practically, that function could align cross-linguistic patterns for the corresponding tokens (ru <-> en) and be applied weight coefficients, depending on the patterns found, to change the weight of distances for certain pairs. 

They demonstrated that for 2 out of 6 directions (ne->en, si->en) the proposed unsupervised method performed better compared to the supervised model proposed by the competition organizers, and for the remaining 4 it showed comparable results.  In other words, the fully automatic QA method proved to be competitive, and in some cases better, than the non-automatic one.
This approach with empirical selection of weights seems rather contrived. Despite this improvement, it was for the ru->en direction that the model score turned out to be noticeably worse than for the base supervised model.

\section{Our hypotheses}
For our study, we hypothesized the possibility of using the BERTScore to assess the quality translation at the sentence level for en->ru direction, both for machine translation and human translation.

Thus, we divided our hypothesis into two: Machine translations QA and  Human translations QA.
We engaged a linguist-assessor to evaluate the translation quality. Since the task of estimating the continuous value of BERTScore is difficult to formalize for human, it was decided to give only relative rank estimates. The rank correlation coefficients (Spearman-\begin{math}\rho\end{math}, Kendall-\begin{math}\tau\end{math}) were chosen as metrics because these metrics are invariant to any monotonic transformations of the measurement scale. The resulting metrics were calculated as a averages over all test samples.

\section{Data acquisition}
As a data source one of the works by an American writer in the mystical thriller genre was taken. About 100 original sentences were randomly selected from the text.

For Human translations QA these sentences were translated using the following 4 engines: \href{https://www.deepl.com/ru/translator}{DeepL}, \href{https://translate.google.com}{Google Translate}, \href{https://translate.yandex.ru}{Yandex Translate}, \href{https://libretranslate.com}{Libre Translate} 
All versions were ranked from 1 to 4 by assessor such that a higher rank corresponds to a better translation quality. Samples with uncertain rank were removed. Finally, we got a dataset of 70 samples.

For Human translations QA, we took 2 translations made by different humans and 1 machine’s (\href{https://translate.google.com}{Google Translate}), all these versions were ranked from 1 to 3 by assessor. 
Samples with uncertain rank were removed as well, and we got dataset of 58 samples.

To exclude bias in the evaluation in both stages, the assessor was not provided with information on how a particular version of the translation was obtained.

\section{Method}

\subsection{Selection of pre-trained contextual models and solving the mismatching issue}
To implement the accurate calculation of BERTScore it is crucial to obtain a common contextual embedding space for tokens in the original English sentences and for tokens of correspondent sentences in Russian.

We tried two approaches:
\begin{itemize}
\item Using a Multilingual model. \cite{Lei_Zhou-2020} followed this path in their work and encountered the “mismatching issue”. Here we used \href{https://huggingface.co/bert-base-Multilingual-cased}{BERT Multilingual base model}.
\item Using 2 Monolingual models \cite{Tomas_Mikolov-2013,Chao_Xing-2015}. In this case, it is necessary to perform the so-called cross-lingual word alignment for two embeddings. Here we used \href{https://huggingface.co/bert-base-cased}{BERT base model (cased)} and \href{https://huggingface.co/DeepPavlov/rubert-base-cased}{DeepPavlov/rubert-base-cased}.
\end{itemize}

There was at least 2 works there authors aligned the BERT contextual embeddings and exploited the idea to learn a transformation in the anchor space. \cite{Tal_Schuster-2019} learned transformation on averaged contextualized embeddings, \cite{Yuxuan_Wang-2019} learned transformation directly in the contextual space. We followed the last one to preserve word semantics as much as possible.

In case of the using separate Monolingual models for 2 languages, the mismatching issue is less apparent, since the tokenizer in each of the models works only with its own language and the division into tokens is more granular as every token is most often a full word, which even more noticeable for the Russian language with its rich morphology.

For example, let's look at the sentence in Russian and results of WordPiece-tokenization with 2 models:
\begin{flushleft}
\selectlanguage{russian}
Вдруг что-то выпало оттуда — большой, неровно сложенный кусок коричневой бумаги.
\item [\textbf{BERT Multilingual tokinizer:}] 'В', '\#\#дру', '\#\#г', 'что', '-', 'то', 'вы', '\#\#пал', '\#\#о', 'от', '\#\#ту', '\#\#да', '[UNK]', 'большой', ',', 'не', '\#\#ров', '\#\#но', 'сл', '\#\#ожен', '\#\#ный', 'к', '\#\#ус', '\#\#ок', 'кор', '\#\#ичне', '\#\#вой', 'бу', '\#\#ма', '\#\#ги', '.'
\item [\textbf{RuBERT Monolingual tokinizer:}] 'Вдруг', 'что', '-', 'то', 'выпало', 'оттуда', '—', 'большой', ',', 'неров', '\#\#но', 'сложен', '\#\#ный', 'кусок', 'коричневой', 'бумаги', '.'
\end{flushleft}
\selectlanguage{british}

This example shows that for Multilingual model almost all the words after tokenizer are spitted into 2-3 tokens, as a result, semantics of words is lost and the mismatching issue will be serious. On the contrary, for Monolingual model the only small number of words are splitted into separate tokens, and semantics of most words is not violated, so mismatching issue will not be serious. 
\begin{flushleft}

To improve matching, we did a simple trick: combined all incomplete WorkPiece - tokens into meaningful words:
\selectlanguage{russian}
\item [\textbf{BERT Multilingual tokinizer:}] 'В', '\#\#дру', '\#\#г' -> 'Вдруг'; 'вы', '\#\#пал', '\#\#о' -> 'выпало'; 'от', '\#\#ту', '\#\#да' -> 'оттуда'; 'не', '\#\#ров', '\#\#но' -> 'неровно'; 'сл', '\#\#ожен', '\#\#ный' -> 'сложенный'; 'к', '\#\#ус', '\#\#ок' -> 'кусок'; 'не', '\#\#ров', '\#\#но' -> 'неровно’; 'сложен', '\#\#ный' -> 'сложенный'; 'кор', '\#\#ичне', '\#\#вой' -> 'коричневой'; 'бу', '\#\#ма', '\#\#ги' -> 'бумаги'
\item[\textbf{RuBERT Monolingual tokinizer:}] 'неров', '\#\#но' -> 'неровно’; 'сложен', '\#\#ный' -> 'сложенный’
\end{flushleft}
and to obtain the final vectors, we averaged the original vectors.

\subsection{Cross-lingual word alignment}
% \label{sect:pdf}

In order to align two embeddings derived from the two pre-trained Monolingual BERT models it is necessary to perform cross-lingual word alignment, which is based on idea that all common languages share concepts that are grounded in the real world \cite{Tomas_Mikolov-2013}.

The conditions for this alignment can be formulated as follows:
\begin{itemize}
\item minimizing the angles between word vectors in two different languages with the same meaning to align word semantics
\item preserving the angles between the word vectors of the original space to save the word relative semantics
\end{itemize}

This operation is reduced as orthogonal Procrustes problem \cite{Gower_1975}: for a given matrix $A$ and matrix $B$ one is asked to find an orthogonal transformation matrix $\Omega$ that minimizes the distance $\Omega{A}$ to $B$:

\begin{equation} 
\underbrace{argmin}_{\Omega^{T}\Omega={I}}\|\Omega{A}-B\|_F
\end{equation}
, where \begin{math}\|.\|_F\end{math} denotes the Frobenius norm. 

This problem has an exact solution found in 1964 \cite{Sch_1966}, it uses the operation of singular matrix decomposition. 

For our case the ru-embedding is treated as matrix $A$ and the corresponding en-embedding is treated as matrix $B$. The orthogonality is necessary to ensure that the angles between vectors of the original embedding are not distorted during the transformation. To find $\Omega$, which minimizes the angles between vectors of words with the same meaning in two different languages after the alignment it is necessary to select such word pairs with the same meaning and solve Procrustes problem for them. These word pairs are called anchors, and the found $\Omega$ optimizes this transformation for arbitrary ru-word vectors.

Thus, the following operations were performed:
\begin{itemize}
\item For all the sentences and for each version of theirs translation, all ru-words were translated using \href{https://translate.google.com}{Google Translate} (a simple frequency dictionary could also be used).
\item The found word translations were compared with the corresponding en-words in the original sentences for full match, in the case of repetition of words in sentences, their sequence was also taken into account, thus identifying the anchor pairs for which transformation is required
\item For the anchor pairs, found in the previous step, A and B matrices were found by concatenating the corresponding ru-word vectors and en-vectors
\item Orthogonal matrix $\Omega$ was found, which realizes $A\rightarrow{B}$ transformation
\item $\Omega$ was used to transform all ru-words vectors of the sentence to the target embedding (optional)
\end{itemize}

\begin{figure}[h]
  \centering
  \includegraphics[angle = 270, width=16cm]{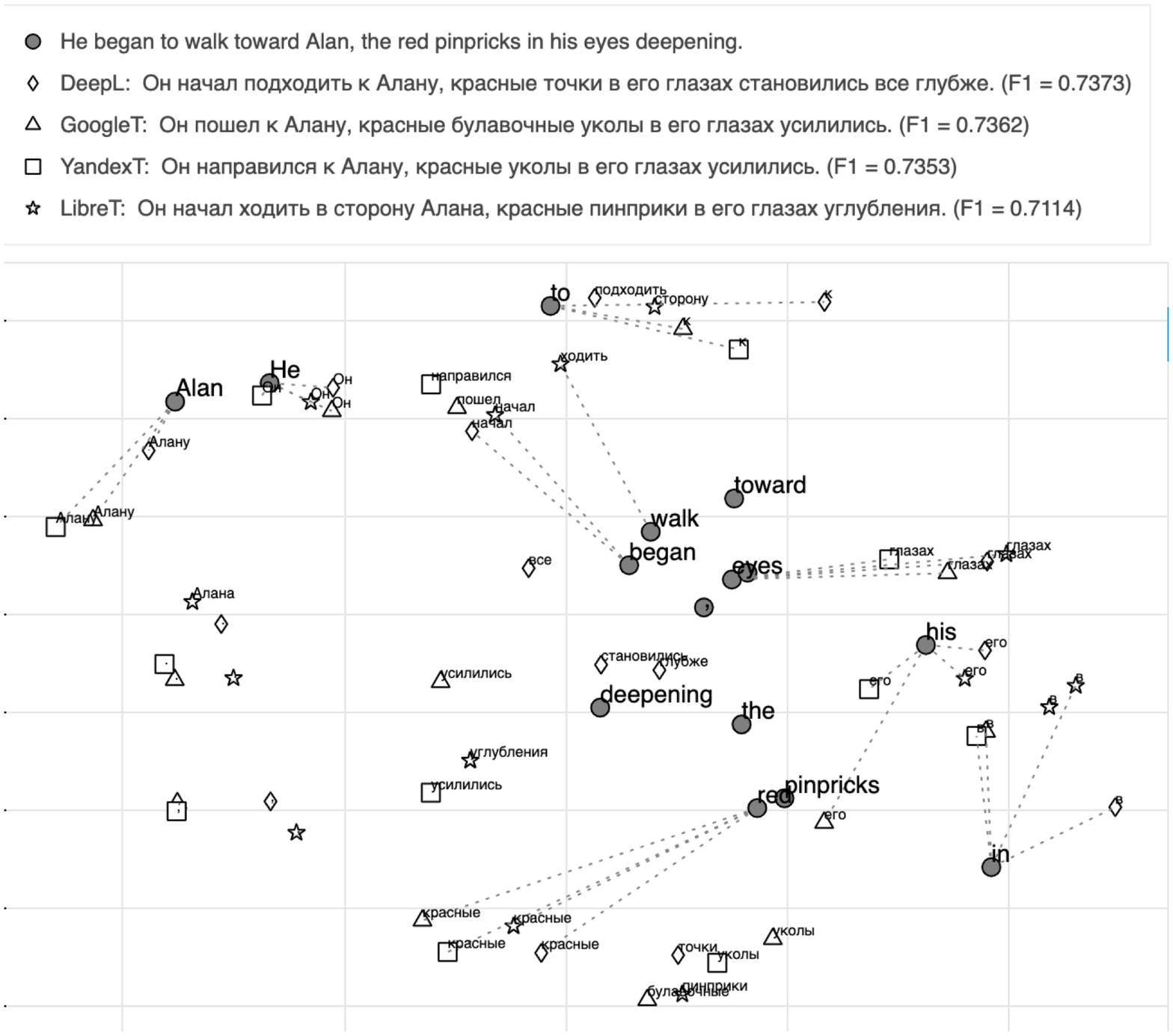}
  \caption{Common contextual word embedding visualization for one of the sentences and its 4 translations after cross-lingual alignment. UMAP dimensional reduction technique is used. Dashed lines indicate anchor pairs. BERTScore F1 for all translations is given in brackets in the legend.}
  \label{emb}
\end{figure}

As a result of the above operations, a single agreed ru-en contextual embedding of the source sentence's words and its translations was obtained for all sentences. Figure~\ref{emb} illustrates this of result for one of the sentences.

\subsection{BERTScore calculation}
BERTScoreF1 was calculated using two methods:
\begin{itemize}
\item With all tokens in sentence, as it was in original work \cite{Tianyi_Zhang-2019},
\item With anchor tokens only. This modification simplifies and accelerates the calculation, reduces mismatching issue and, as it turned out, finally showed the best results.
\end{itemize}

Spearman-\begin{math}\rho\end{math} and Kendall-\begin{math}\tau\end{math} were calculated for each test sample, and finally the results were averaged over all samples.

\section{Results}

Multilingual BERT embedding showed poor results both for Machine and Human translation, as the rank correlations for all the approaches are less then zero (Table~\ref{QA}). The most likely reason is mismatching issue. Further we will focus on the results of 2 Monolingual embedding.
\begin{table}[h]
% \vspace{-1cm}
\begin{center}
\begin{tabular}{|l|c|c|}
\hline \bf Machine translation QA & \bf Spearman & \bf Kendall \\ \hline
Multilingual (all tokens) & -0.071 & -0.061 \\
Multilingual (anchors only) & -0.037 & -0.023 \\
2 Monolingual + alignment (all tokens) & 0.329 & 0.272 \\
2 Monolingual + alignment (anchors only) & \bf0.571 & \bf0.476 \\
% \hline
\hline \bf Human translation QA & \bf Spearman & \bf Kendall \\ \hline
Multilingual (all tokens) & -0.431 & -0.355 \\
Multilingual (anchors only) & -0.297 & -0.246 \\
2 Monolingual + alignment (all tokens) & -0.352 & -0.283 \\
2 Monolingual + alignment (anchors only) & \bf0.102 & \bf0.061 \\
\hline
\end{tabular}
\end{center}
  \caption{Average of Spearman-\begin{math}\rho\end{math} and Kendall-\begin{math}\tau\end{math} rank correlation coefficients between BERTScoreF1 and human judgments for Machine and Human translation QA, 2 types of embeddings}
\label{QA}
\end{table}

\begin{figure}[h]
  \centering
  \includegraphics[angle = 0, trim = 5cm 8.5cm 5cm 8.5cm, width=7cm]{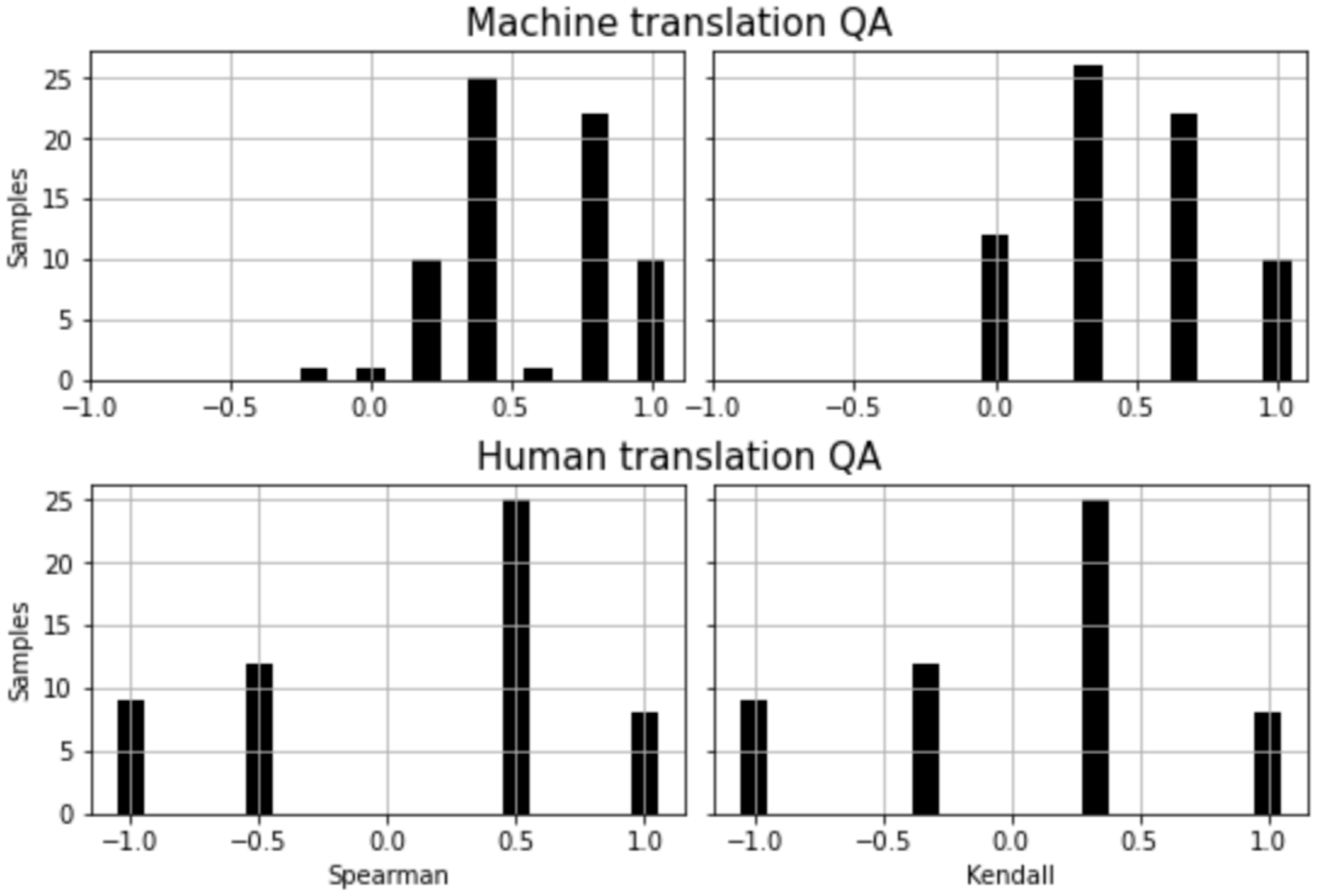}
  \caption{Spearman-\begin{math}\rho\end{math} and Kendall-\begin{math}\tau\end{math} rank correlation coefficients between BERTScoreF1 and human judgments for Machine and Human translation QA. 2-Monolingual embeddings with word alignment approach.}
  \label{mht}
\end{figure}

2-Monolingual embeddings with word alignment: the average values of rank correlation coefficients are strictly above zero for Machine translation QA (Table~\ref{QA}). Figure~\ref{mht} also indicates that for most the test samples BERTScoreF1 correlates well with human judgments. So, our first hypothesis that BERTScore can be used to evaluate machine translation is confirmed for this method.

The second hypothesis about Human translation QA was not confirmed, since the values of the rank correlations for all approaches are close to zero (Table~\ref{QA}), which roughly corresponds to a random guess. The same is confirmed by the Figure~\ref{mht} of the distribution of the rank correlation values for all the samples. To clarify the reasons for this result, we analyzed several cases where BERTScore results were opposite to expert estimates of the algorithm's performance (Spearman-\begin{math}\rho = -1\end{math}).
\begin{flushleft}
Let’s have a look on one of them.
\item [\textbf{Initial sentence:}] 'He had learned that lesson yesterday when he had come home to find those terrible pink slips taped up all over the house'

and 3 version of translation:
\selectlanguage{russian}
\item [\textbf{Human translation 1:}] 'Вчера он получил наглядное тому подтверждение, когда вернулся домой и обнаружил эти страшные розовые листки, расклеенные по всему дому.' BERTScoreF1 = 0.7252
\item [\textbf{Human translation 2:}] 'Он понял это, когда, вернувшись вчера домой, обнаружил эти жуткие розовые талоны, расклеенные повсюду.' BERTScoreF1 = 0.7574
\item [\textbf{\href{https://translate.google.com}{Google Translate}:}] 'Он усвоил этот урок вчера, когда, придя домой, обнаружил эти ужасные розовые бланки, заклеенные по всему дому.' BERTScoreF1 = 0.7620
\end{flushleft}

The BERTScore seems to be higher for the Machine version  (\href{https://translate.google.com}{Google Translate}) and lower for the Human versions due to closer in both word count and word semantic. 
This example indicates the following reasons for poor correlation:
\begin{itemize}
\item In human versions of translations, semantic proximity to the original sentence is much harder to establish than in machine translations, because these versions include paraphrasing, which is perceived as "artistic language" and is judged higher by humans, despite the greater distance in semantics.
\item The assessor appears to be more likely to better score a version that considers the broader context beyond a one sentence, as opposed to the BERT, which considers the context of no more than one sentence.
\end{itemize}
For the both assessments, the anchors-only method performed noticeably better than the all-words based approach.
\section{Conclusion and next steps}
Based on the results, we can conclude that BERTScore is suitable for automatic Machine translations QA on sentences level, but not suitable for automatic Human translations QA.
The most prominent results were obtained with the following approaches:
\begin{itemize}
\item Using pair of pretrained Monolingual BERT models without fine-tuning,
\item Using word alignment based on anchor words, which is obtained by word translation
\item Combining all incomplete WorkPiece tokens into meaningful words and using simple averaging of corresponding vectors,
\item BERTScore calculation on anchor words only.
\end{itemize}

In the future, we plan to perform additional testing for the en->ru direction on publicly available annotated corpora.

For translation directions with linguistically similar language pairs, the mismatching issue will not be so evident due to the effect of big vocabulary overlap\cite{Telmo_Pires-2019}. It can be expected that the same approach for calculating BERTScore can show better results for such languages with both Monolingual BERT and Multilingual BERT.

\section*{Acknowledgements}
We thank Irakli Menteshashvili for his help in assessing translations and working with literary sources.

\bibliography{approach.bib}
\bibliographystyle{dialogue}

%\begin{thebibliography}{}

%\end{thebibliography}

\end{document}